\documentclass[letterpaper, 10 pt, conference]{ieeeconf}  

\IEEEoverridecommandlockouts

\usepackage{graphicx} 
\usepackage{amsmath} 
\usepackage{amssymb}  

\usepackage{amsthm}
\usepackage{soul}
\usepackage[dvipsnames]{xcolor}
\usepackage[export]{adjustbox}

\usepackage{gensymb}

\theoremstyle{definition}

\newtheorem{problem}{Problem}
\newtheorem{remark}{Remark}

\newtheorem{example}{Example}
\usepackage{bm}
\usepackage{wrapfig}
\usepackage{cuted}
\usepackage{subcaption}

\usepackage{caption}
\usepackage{dsfont}
\DeclareCaptionType{equ}[][]
\graphicspath{{figures/}} 
\usepackage{hyperref}
\hypersetup{
    colorlinks=true,
    linkcolor=blue,
    filecolor=magenta,      
    urlcolor=cyan,
    pdftitle={Overleaf Example},
    pdfpagemode=FullScreen,
    }
\usepackage{algorithm}  
\usepackage{algpseudocode}[noend]
\graphicspath{{figures/}}

\DeclareMathOperator*{\argmin}{arg\,min}
\newcommand{\reals}{\mathbb{R}}
\newcommand{\s}{\mathcal{S}}

\newcommand{\eqn}[1]{\begin{align}#1\end{align}}

\usepackage{ulem}

\title{ \LARGE \bf Risk-aware MPPI for Stochastic Hybrid Systems}

\author{Hardik Parwana$^{1}$, Mitchell Black$^{2}$, Bardh Hoxha$^{3}$, Hideki Okamoto$^{3}$, Georgios Fainekos$^{3}$, \\Danil Prokhorov$^{3}$, Dimitra Panagou$^{1,4}$ 
    \thanks{$^{1}$ Department of Robotics, $^{4}$ Department of Aerospace Engineering, University of Michigan, Ann Arbor, USA {\tt\small \{hardiksp, dpanagou\}@umich.edu}
    }
    \thanks{$^{2}$ MIT Lincoln Laboratory, Lexington, MA 02421, USA {\tt\small mitchell.black@ll.mit.edu}. This work was started when Mitchell Black was with TMNA.}
 \thanks{$^{3}$ Toyota Motor North America (TMNA), Research \& Development, Ann Arbor, MI 48105, USA
 {\tt\small {<first\_name.last\_name> }@toyota.com.}
 }
}

\begin{document}

\maketitle
\thispagestyle{empty}
\pagestyle{empty}

\begin{abstract}
     Path Planning for stochastic hybrid systems presents a unique challenge of predicting distributions of future states subject to a state-dependent dynamics switching function. 
     In this work, we propose a variant of Model Predictive Path Integral Control (MPPI) to plan 
     kinodynamic paths for such systems. 
     Monte Carlo may be inaccurate when few samples are chosen to predict future states under state-dependent disturbances. 
     We employ recently proposed Unscented Transform-based methods to capture stochasticity in the states as well as the state-dependent switching surfaces. 
     This is in contrast to previous works that perform switching based only on the mean of predicted states. 
     We focus our motion planning application on the navigation of a mobile robot in the presence of dynamically moving agents whose responses are based on sensor-constrained attention zones.
     We evaluate our framework on a simulated mobile robot and show faster convergence to a goal without collisions when the robot exploits the hybrid human dynamics versus when it does not.
     
\end{abstract}


\section{Introduction}

Hybrid dynamical systems \cite{lygeros2003dynamical, goebel2009hybrid}, in which the dynamics function changes in response to an \textit{event}, appear in several robotic applications. 
Examples include walking legged robots~\cite{kong2023hybrid}, contact-rich quadrotor navigation~\cite{kong2021ilqr}, and mobile robot navigation on different terrains~\cite{d2023stochastic}. 
In practice, due to disturbances or \textit{a priori} unknown dynamics of the robot or the environment, these systems are often modeled as stochastic hybrid systems~\cite{alwan2016recent}. 
While there is extensive research on path planning and control for hybrid dynamical systems~\cite{buss2002nonlinear,li2020hybrid}, we focus here on model predictive approaches to address these challenges.

Our motivation for studying these systems arises from navigation challenges, 
and particularly two interrelated, but often independently addressed, challenges.
First, the motion of agents, whether human or non-human, is challenging due to noisy observations of past states and the often unpredictable behavior of these agents (e.g., sudden changes in direction). Planners must account for this inherent uncertainty, which may be captured by modeling agent motion as a stochastic hybrid system.
Second, agents may abruptly alter their behavior upon perceiving new entities in their environment, including robots, obstacles, or other agents. In this work, we model agents’ dynamic modes as hybrid dynamical systems, where each mode represents varying levels of cooperation or responsiveness to a robot. Mode switching is triggered, for example, when the robot enters an agent’s attention zone, such as a human’s field of view or an aquatic or aerial agent’s sensory range.


The integration of estimators into path planning systems is crucial for navigating environments that involve diverse agents. 
Previous works~\cite{ivanovic2019trajectron, vemula2018social} have proposed estimators that provide deterministic predictions, typically trained on datasets specific to constrained contexts like pedestrian traffic in mall corridors. More recent research has shifted towards developing adaptable behavioral models for estimation~\cite{thakkar2023adaptive, bera2019emotionally, minami2023development} or inferring deviations from expected motion based on observations~\cite{tian2022safety}. 
These approaches consider factors such as social attention, which prioritizes nearby agents~\cite{vemula2018social}, or context-specific parameters like environmental constraints and sensory limitations~\cite{thakkar2023adaptive}.
In this work, we focus on modeling cooperative behaviors where agent interactions are influenced by sensory zones, such as a field of view or sensory range.
These factors play a critical role in enhancing situational awareness and guiding responses to dynamic environmental changes~\cite{minami2023development}.

Since exact analytical expressions for propagating probability distributions through nonlinear functions are generally not available, several approximate methods have been developed to solve stochastic MPC. These methods include Monte Carlo simulations~\cite{amadio2022model}, successive Gaussian approximation~\cite{deisenroth2011pilco}, and the Unscented Transform~\cite{parwana2022foresee}. In the domain of hybrid systems, MPC problems are typically formulated as mixed-integer optimizations~\cite{d2023stochastic, bemporad2010model}, which can become computationally prohibitive, particularly when the system dynamics are nonlinear. 

Since solving the hybrid MPC exactly may not be feasible for real-time implementation, we adopt Model Predictive Path Integral (MPPI) Control~\cite{williams2018information}, a sampling-based, gradient-free method for solving unconstrained MPC problems in real-time. 
MPPI typically converts hard constraints into soft constraints by incorporating them into the objective function. 
Variants of MPPI have been developed to address uncertainty in the initial state~\cite{mohamed2023towards} and disturbances in system dynamics~\cite{williams2018robust, gandhi2021robust}. However, \cite{mohamed2023towards} focuses on deterministic dynamics, and \cite{williams2018robust, gandhi2021robust} address only state-independent noise.
To our knowledge, applying MPPI to systems with state-dependent disturbances remains a challenge due to the lack of analytical methods for propagating future states under such disturbances. 
In this work, we leverage recent advances in uncertainty propagation techniques\cite{parwana2022foresee} for stochastic MPC within the MPPI framework.


The problem of switching dynamics under uncertain states is similar to work on belief tree planning for motion planning under estimator uncertainty~\cite{bry2011rapidly}. These works incorporate state estimation uncertainty in the planning module to probabilistically bound the safety violation metric as the robot navigates with an onboard localization module. The correction step of the Extended Kalman filter (EKF) that is used for state estimation is triggered only when an obstacle is within the sensing region of the robot. However, \cite{bry2011rapidly} implements the switch between these different modes based only on the mean of the state estimate, thereby making switching a deterministic phenomenon despite the state estimates being stochastic. 

In this work, we propose to use a particle-based uncertainty propagation scheme based on Unscented Transform~\cite{parwana2022foresee} in MPPI. 
{\bf Our contributions are twofold:} 
First, by utilizing weighted particles, we map the stochasticity of states to the stochasticity of mode switching, moving beyond deterministic switching based only on predicted state means. 
Second, we extend MPPI to systems with arbitrary state-dependent disturbances. Both of these contributions enable us to define a risk-aware framework for path planning with respect to static and dynamic obstacles for a hybrid dynamical system.
Our experimental results demonstrate improved cost and safety performance 
by accounting for the hybrid nature of non-ego agent dynamics, compared to planning approaches that neglect cooperative behaviors or assume constant velocity models~\cite{vulcano2022safe, black2023future}.



\subsection{Related Work}

The work most similar to ours in terms of solution approach is ~\cite{cai2023probabilistic}, wherein different terrains induce different dynamics for a mobile robot. 
An unknown traction parameter in the dynamics is modeled with a terrain-dependent probability distribution, thereby making this also an instance of stochastic hybrid dynamics. 
They propose a Monte Carlo (MC) scheme motivated by ~\cite{wang2021adaptive} to evaluate the cost of executing a control input sequence. 
From an application standpoint, \cite{cai2023probabilistic} applied MPPI when the uncertainty in dynamics is dependent on static terrain features whereas uncertainty in our case stems from active interaction between dynamically moving ego and non-ego agents.
However, the main difference is that the method presented in~\cite{cai2023probabilistic} uses Monte Carlo whereas we use UT to propagate state distributions across nonlinear functions. 
Whereas MC can require a prohibitively large number of samples~\cite{ghavamzadeh2006bayesian}, for similar performance in accuracy UT is shown to be more computationally efficient in contrast~\cite{ebeigbe2021generalized, angrisani2005unscented, parwana2022foresee}. 








\section{Notation and Problem Formulation}
The set of real numbers is denoted as $\reals$. For $x\in \reals, y\in \reals^n$, $|x|$ denotes the absolute value of $x$ and $\|y\|$ denotes the $L_2$ norm of $y$. The time derivative of $x$ is denoted by $\dot x$ and $\frac{\partial F}{\partial x}$ denotes the gradient of a differentiable function $F: \reals^n \rightarrow \reals$ with respect to $x\in \reals^n$.

Consider a robot with state $x_r \in \reals^{n_r}$, control input $u_r\in \reals^{m_r}$, and dynamics
\eqn{
x_{r,t+1} = f_r(x_{r,t}, u_{r,t}).
\label{eq::robot_dynamics}
}
Each agent $p$ is modeled as a stochastic hybrid (i.e., multimodal) dynamical system with continuous state $x_{p} \in \reals^{n_p}$ and $N$ operating modes $q_j$, for $j\in \{1,2,\ldots,N\}$. 
Each mode corresponds to different agent dynamics. For example, a robot may exhibit different behavior toward another agent when it enters its limited sensing region.

When the agent $p$ is in mode $q_j$, then their motion is governed by stochastic dynamics $D_{q_j}:\reals^{n_p}\times \reals^{n_r}\rightarrow \reals^{n_p}$, which depend both on the state of the agent and the robot, that is,
\eqn{
x_{p,t+1} \sim D_{q_j}(x_{p,t}, x_{r,t}).
\label{eq::human_stochastic_dynamics}
}
A mode $q_j$ is active in a region 
\begin{gather*}
  \mathcal{A}_j=\{ (x_p, x_r) \in \reals^{n_p}\times \reals^{n_r} ~|~ \wedge_{k \in \bar K} g_{j,k}(x_p, x_r)\leq 0 \},
\end{gather*}
where $g_{j,k}:\reals^{n_p}\times \reals^{n_r}\rightarrow \reals, k\in \bar K=\{1,\dots,K\}$ are $K$ activation functions for mode $q_j$.
We assume that the regions $\{ \mathcal{A}_j \}_{j\leq N} $ form a partition of the space $\reals^{n_p}\times \reals^{n_r}$.

\begin{remark}
To simplify the presentation and notation, we have introduced a hybrid dynamical model where the activation functions partition the agent-robot state space.
However, the motion planning framework that we introduce in this paper is more general and can handle general hybrid dynamical systems \cite{Mitra2021book}.
\end{remark}

In our framework, we allow for multiple dynamic agents. 
When necessary, we will use the subscript $i$, e.g., $x_{p,i}$, $D_{i,q_j}$, etc, to highlight the fact that multiple agents are present -- potentially under different stochastic dynamics. 

\begin{example}[Simple motion model with attention radius]
    Consider the following attention regions corresponding to two operating modes $q_{1}$ (non-attentive) and $q_{2}$ (attentive) of an external agent with corresponding regions
    \begin{align*}
    \mathcal A_{1} & = \{ (x_p, x_r) ~|~ ||x_p - x_r|| >  d_s \} \\
    \mathcal A_{2} & = \{ (x_p, x_r) ~|~ ||x_p - x_r||\leq  d_s \} 
    \end{align*}
    where $d_s$ is the sensing area of the agent. 
    The stochastic dynamics in each mode are: 
    \begin{align*}
        D_{q_1}(x_{p,t},x_{r,t}) & = x_{p,t} + u_{g}(x_{p,t}) + w_{1,t}(x_{p,t})  \\
        D_{q_2}(x_{p,t},x_{r,t}) & = x_{p,t} + u_{g}(x_{p,t}) + u_{a}(x_{p,t}, x_{r,t}) + w_{2,t}(x_{p,t}) 
    \end{align*}
    where $x_{p,t}$ and $x_{r,t}$ are the positions of the agent and the robot, respectively, at time $t$, 
    $u_{g}$ models the agents's attraction toward their goal, 
    $u_{a}$ models the agent's avoidance of the ego-robot, and 
    $w_{j,t}$ are appropriate noise terms that may depend on the state of the agent in general.
\end{example}

In general, the agents dynamics can also be written as
\eqn{
  x_{p,t+1} \sim \sum_{j=1}^N \delta_j D_{q_j}(x_{p,t}, x_{r,t})
}
where $\delta_j$ are binary variables and $\delta_j=1$ if mode $q_j$ is active,
\eqn{
\delta_j = \left\{ \begin{array}{cc}
   1  & \mbox{ if } (x_{p,t}, x_{r,t}) \in \mathcal{A}_j \\
   0 & \mbox{ otherwise }
\end{array}
\right. 
\label{eq:switching}
}

\subsection{Safety Constraints}
Our objective is to navigate a mobile robot in an environment with the presence of static obstacles, such as walls, and stochastic, dynamic agents. 
We model the requirement for collision avoidance between the robot $x_r$ and each agent $x_p$ and between the robot $x_r$ and each static obstacle $x_o$ using distance functions:  
\begin{subequations}
    \begin{align}
  h_p : \reals^{n_r} \times \reals^{n_p} \rightarrow \reals^+ \label{eq::constraint_function}\\
  h_o : \reals^{n_r} \times \reals^{n_o} \rightarrow \reals^+
\end{align}
\end{subequations}
We further assume that $h_p$ and $h_o$ are defined in such a way that given a safety threshold $\varepsilon>0$, the sets $\{ (x_r,x_p) \, | \, h_p(x_r,x_p) < \varepsilon \}$ and $\{ (x_r,x_o) \, | \, h_o(x_r,x_o) < \varepsilon \}$ do not contain disconnected sets of measure zero. 
This means that the safety problem is well-defined.








\subsection{Problem definition}

The question we are answering in this paper is whether a robot can improve its motion planning performance, e.g., time to reach its goal or energy consumption, if the robot utilizes a model of the agents' awareness of their environment. 
At the same time, we would like to ensure that safety is not compromised.
In other words, we aim to improve the system performance while keeping the risk of collisions bounded by a desired threshold $\varepsilon$.

Formally, we aim to solve the following path planning problem:

\begin{problem}
\label{problem::constant_risk}
    Given the system comprised of \eqref{eq::robot_dynamics} and \eqref{eq::human_stochastic_dynamics}, 
    a safety threshold $\varepsilon>0$,
    a risk threshold $\epsilon \in [0,1]$, and 
    a stage-wise cost $R:\reals^{n_r}\times \reals^{m_r}\rightarrow \reals$,
    plan a control input trajectory $v = \{u_{r,1}, u_{r,2}, ..., u_{r,T}\}$ for a horizon $T$ that solves the following optimization problem
    \begin{subequations}
        \eqn{
    v = \argmin_{\{u_{r,1}, ..., u_{r,T}\}} \quad \sum_{\tau=0}^T R(x_{r,\tau}, &u_{r,\tau}) \\
    \textrm{s.t.} \quad x_{r,\tau+1} = f_r(x_{r,\tau}, &u_{r,\tau}) \\
                 x_{p,\tau+1} \sim \sum_{i=1}^N \delta_{j,\tau} &D_{q_j}( x_{p,\tau}, x_{r,\tau} ) \label{eq::problem_dynamics}\\
                 \delta_{j,\tau} = 1\textrm{ if } (x_{p,\tau}, &x_{r,\tau}) \in \mathcal{A}_j \\
                 P\left[ \bigwedge_{k=\tau}^T h_p(x_{r,k}, x_{p,k}) \geq \varepsilon \right] &\geq 1 - \epsilon \label{eq::chance_constraint}\\
                 P\left[ \bigwedge_{k=\tau}^T h_o(x_{r,k}, x_{o,k}) \geq \varepsilon \right] &\geq 1 - \epsilon
    }
    \end{subequations}
\end{problem}

    


\section{Preliminaries}
Instead of solving Problem \ref{problem::constant_risk} exactly, we solve it approximately by imposing hard constraints as penalties in the objective function
and using MPPI to find a locally optimal path. In this section, we briefly describe the MPPI algorithm as well as the Expansion-Compression UT algorithms that help approximate Eq. \eqref{eq::problem_dynamics}. 

\subsubsection{Model Predictive Path Integral Control}

MPPI finds a locally optimal solution to an unconstrained optimal control problem by sampling trajectories around a nominal trajectory and assigning weights to the trajectories based on the cost incurred. The final trajectory results from a weighted summation of all the sampled trajectories. 

Consider a nonlinear system with state $x_t \in \reals^{n}$ and control input $u_t \in \reals^m$ that follows the following discrete-time dynamics
\eqn{
   x_{t+1} = F(x_t, u_t).
}
For a time horizon $H$, consider the state trajectory $\mathbf{x}=[x_t^T, ..., x_{t+H}^T]^T$, mean control input sequence $\mathbf{v} = [v_t^T, .., v_{t+H}^T]^T, v_\tau\in \reals^m$ and injected Gaussian noise $\mathbf{g} = [g_t^T,..,g_{t+H}^T]^T$ where $g_\tau \sim N(0,\Sigma_g)$. Let the disturbed control input sequence be $\mathbf{u}=\mathbf{v}+\bm{g}$. 

The MPPI algorithm solves the following problem
\begin{subequations}
    \eqn{
   \min_{\mathbf{v}} \quad  & J(\mathbf{v}) = \mathbb{E} \left[ \Phi(x_{t+H}) + \sum_{\tau=t}^{t+H-1} \left( Q(x_\tau) + \frac{\lambda}{2} v_\tau^T \Sigma_g^{-1} v_\tau \right)  \right] \label{eq::mppi_objective_org}\\
   \textrm{s.t.} \quad & x_{\tau+1} = F(x_\tau, v_\tau+g_\tau) \\
   & g \sim \mathcal{N}(0,\Sigma_g) \label{eq::MPPI-disturbance}
}
\end{subequations}
Below we provide a summary of the MPPI algorithm and refer the reader to \cite{williams2018information} for a more theoretical insight into its derivation and optimality properties.
MPPI samples $M$ trajectories where $M\gg 1$, and each trajectory is affected by a different disturbance sequence $\epsilon$. We will add a superscript $()^m$ to denote quantities corresponding to $m^{th}$ sampled trajectory. The cost $S_m$ of each trajectory $m\in \{1,..,M\}$ is evaluated based on \eqref{eq::mppi_objective_org} as follows
\eqn{
S_m  = \Phi(x_{t+H}^m) \sum_{\tau=t}^{t+H-1}Q(x_\tau^m) + \gamma (v_\tau^m)^T \Sigma_g^{-1} (v_\tau^m + g_\tau^m)
\nonumber 
}
where $\gamma\in [0,\lambda]$. The weight $w_m$ for $m^{th}$ trajectory is determined as
\eqn{
w_m = \exp \left( -\frac{1}{\lambda} (S_m - \beta)  \right)
\label{eq::mppi_weight}
}
where $\beta=\min_{m\in\{1,..,M\}}S_m$. The optimal control sequence is then computed as
\eqn{
\mathbf{v}^+ = \sum_{m=1}^M w_m \mathbf{u}^m / \sum_{m=1}^{M}w_m.
\label{eq::mppi_final_control}
}
In a receding horizon implementation, the control input is chosen as $u_t=\mathbf{v}^+_0$ and the mean sequence $\mathbf{u}$ for the next iteration is set to be $\mathbf{v}=\mathbf{v}^+$. 

\subsubsection{Expansion - Compression Unscented Transform}
Given an initial state, $x_{p,t}$, predicting future state distributions under stochastic dynamics such as \eqref{eq::human_stochastic_dynamics} requires successively evaluating the following integral for all $t\in \{1,..,T\}$
\eqn{
  p(x_{p,t+1}) = \int p( x_{p,t+1} | x_{p,t}; D_{q_i} )p(x_{p,t}) \mathrm{d} x_{p,t}.
  \label{eq::state_prediction}
}
Several approximate numerical procedures exist to evaluate \eqref{eq::state_prediction} as it cannot be evaluated analytically in general. In this work, we employ Expansion-Compression Unscented Transform \cite{parwana2022foresee} (ECUT), a method that exploits Unscented Transform (UT) to propagate state distributions across state-dependent uncertain functions. 

At time $t$, for $x_{p,t}\in\mathbb{R}^n$, consider a set $\s_t$ of $N$ particles, or sigma points, $\s_{t,i}\in \reals^{n_h}$, $i=1,\dots,N$, denoted $\s_t=\{\s_{t,1},\s_{t,2},\dots,\s_{t,N}\}$, and the set $w_t=\{w_{t,1},w_{t,2},\dots,w_{t, N}\}$ comprising their associated weights $w_{t,i}$. The sample (or empirical) mean and covariance of $x_{p,t}$ are computed as follows:
\begin{subequations}
\begin{align}
    \mathbb{E}_s[x_{p,t}] &= \sum_{i=1}^N w_{t,i}\s_{t,i}, \\ 
    \Sigma_s[x_{p,t}] &=  \sum_{i=1}^{N} w_{t,i} (\s_{t,i}-\mathbb{E}_s[x_{p,t}])(\s_{t,i}-\mathbb{E}_s[x_{p,t}])^T
\end{align}
\label{eq::sigma_point_basic}
\end{subequations}
Given the moments of a distribution of $x_t$, UT provides a principled way of selecting the sigma points $\mathcal{S}_t, w_t$ such that their empirical moments are equal to the true moments of a given distribution. For example, the standard UT algorithm presented in \cite{julier1997new} takes as input the mean and covariance of $n$ dimensional vector and outputs $N=2n+1$ points and weights. The probability distribution of a function of random variables (state vector in our case) is then obtained by passing sigma points through the function. For example, for $x_{t+1}=F(x_t)$, the sigma points $\mathcal{S}_{t+1}$ are obtained as
\eqn{
\mathcal{S}_{t+1,i} = F(\mathcal{S}_{t,i}).
}
When the function $F$ is stochastic, that is $x_{t+1} \sim F(x_t)$, each sigma point $\mathcal{S}_{t,i}$, maps to a distribution of states defined by $F(\mathcal{S}_{t,i})$. This necessitates increasing the number of sigma points to keep track of distributions. That is, if each $\mathcal{S}_{t,i}$ generates $N_F$ points (for example, using vanilla UT algorithm based on knowledge of moments of $F(\mathcal{S}_{t,i}))$, the total number of sigma points for $x_{t+1}$ is $NN_F$. However, this implies an increase in the number of sigma points across multiple time steps and therefore this procedure is not scalable. ECUT\cite{parwana2022foresee} provides a scalable way by augmenting the increase in the number of sigma points at each time step, called Expansion operation, by a moment-matching-based Compression operation. The procedure is presented in Algorithm \ref{algo::ec-ut} for the sake of completeness. The \textit{expand\_sigma\_points} function in Line 4 generates the $NN_F$ sigma points and Lines 5-7 constitute the compression algorithm and generate new $N$ points that have the same mean and covariance as the expanded $NN_F$ points. Note that higher-order moments besides the mean and covariance can also be used to generate sigma points. The interested reader is referred to \cite{parwana2022foresee} for more details.

\begin{algorithm}[h!]
\caption{Expansion-Compression UT}
\begin{algorithmic}[1]
    \Require $(\s_\tau, w_\tau$), $F$ \Comment{current state sigma points, Stochastic Dynamics function}
            \State $\s_{next}, w_{next} \leftarrow $ expand\_sigma\_point($F, \s_{\tau}, w_{\tau}$)
            \State $\mathbb{E}_s = \sum_{i=1}^{NN_F} w_{next,i}\mathcal{S}_{next,i} $
            \State $\Sigma_s = \sum_{i=1}^{N} w_{next,i} (\s_{next,i}-\mathbb{E}_s)(\s_{next,i}-\mathbb{E}_s)^T$
            \State $\s_{\tau+1},w_{\tau+1} \leftarrow \textup{generate\_UT\_points}(\mathbb{E}_s, \Sigma_s)$
\end{algorithmic}
\label{algo::ec-ut}
\end{algorithm}

\section{Methodology: ECUT-MPPI}

We leverage MPPI to propose an approximate solution to \eqref{problem::constant_risk} that is amenable to real-time implementation. 
We make two algorithmic contributions to applying MPPI for stochastic hybrid systems. 
First, we use the Expansion-Compression UT for state prediction under stochastic dynamics with state-dependent distributions. Second, unlike previous works that realize state-dependent switching dynamics based only on the mean on predicted states \cite{bry2011rapidly}, we switch dynamics based on the sigma particles. 
This is essential because although the switching function in \eqref{eq:switching} is a deterministic function of the robot's and agents's position, the uncertainty in the human's position implies that switching at a given time is also probabilistic. 
Using sigma points helps capture this stochasticity in switching. 


We modify the objective of MPPI to be risk-aware as follows. The stage-wise cost is split into two objectives: $Q(x_{r,\tau}) = Q_c(x_{r,\tau}) + Q_h(x_{r,\tau}, x_{p,\tau}, x_o)$, where $Q_c: \mathbb R^{n_r} \mapsto \mathbb R$ is for convergence to the desired robot state $x_{r,d}$, and $Q_h: \mathbb R^{n_r}\times \reals^{n_p} \times \reals^{n_o} \mapsto \mathbb R^+$ is safety, i.e.,
\eqn{
  Q_c(x_\tau) &= \mathbb{E} \left[\gamma_1|| x_r - x_{r,d} ||^2 \right] \label{eq::used_risk} \\
  Q_h(x_\tau) &= \frac{\gamma_2}{ \mu_{h,p} - \alpha \sigma_{h,p} } + \frac{\gamma_3}{ h_{obs} }, \nonumber
}
where $(\mu_{h,p}, \sigma_{h,p})$ are the mean and standard deviation of the smallest distance value \eqref{eq::constraint_function} of distances to all the agents $p$ at time $t$, $h_{obs}$ is similarly the smallest distance of distances to all the obstacles $o$, and $\gamma_1, \gamma_2, \gamma_3>0$ are user-chosen parameters. $\alpha$ is a risk tolerance parameter that can be related to $\epsilon$ in \eqref{eq::chance_constraint}, with larger $\alpha$ promoting lower risk tolerance. Essentially, $\mu_{h,p}-\alpha\sigma_{h,p}$ represents the lower bound of the central confidence interval of a gaussian and, for example, for $\epsilon=0.05$, $\alpha=1.96$.  The MPPI optimization is thus designed as
\begin{subequations}
    \eqn{
   \min_{\mathbf{v}} \quad  & J(\mathbf{v}) = Q_c(x_{t+H}) + \label{eq::mppi_objective}\\
   & \quad \quad  \sum_{\tau=t}^{t+H-1} \left( Q_c(x_\tau) + Q_h(x_\tau) + \frac{\lambda}{2} v_\tau^T \Sigma_\epsilon^{-1} v_\tau \right)\nonumber \\
   \textrm{s.t.} \quad & x_{\tau+1} = F(x_\tau, v_\tau+\epsilon_\tau) \\
   & \epsilon \sim \mathcal{N}(0,\Sigma_\epsilon)
}
\end{subequations}

\begin{figure}[t]
    \centering
    \includegraphics[width=0.50\textwidth]{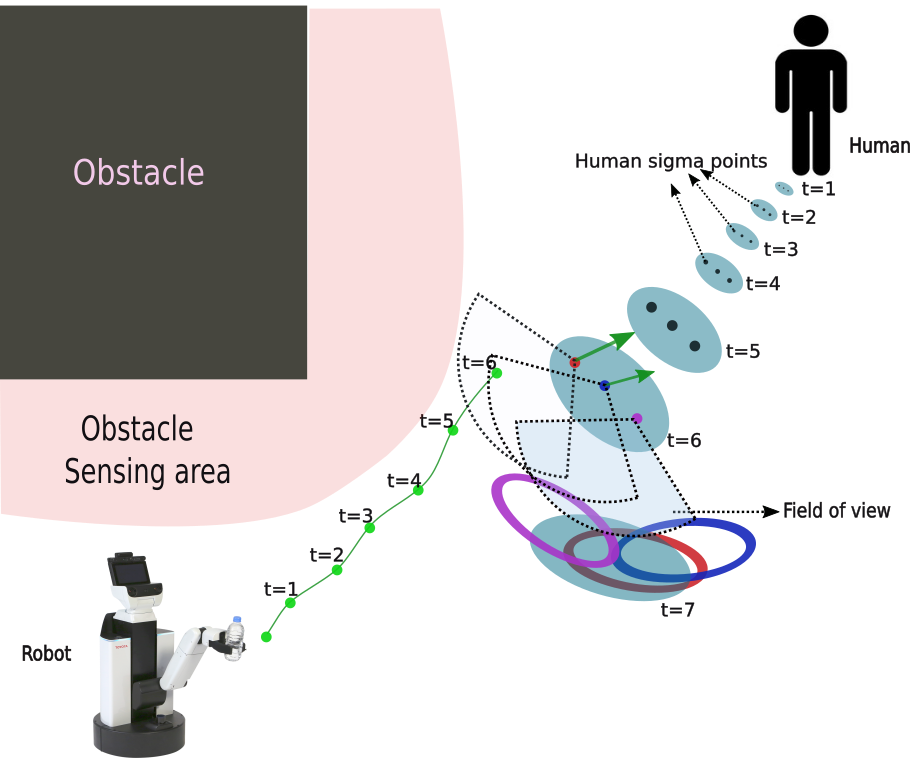}
    \caption{\small{
    The robot's sampled trajectory (green) induces different responses from each of the human's (the non-ego agent in this example) sigma points. The ellipses represent the distribution of predicted human states and the dots inside them represent the sigma points. At $t=6$, the human perceives the robot for the first time in its field-of-view for 2 out of 3 sigma points. The red and blue points get influenced by robot position at $t=6$ while the pink point does not - resulting in three different future ellipses (of the same color respectively). The compression operation combines the three hollow ellipses into a single (solid) ellipse to represent human state distribution at $t=7$. Note that distribution need not be an ellipse and depends on the variant of UT employed.
        }}
    \label{fig::main_fig}
    \vspace{-3mm}
\end{figure}


\subsection{Algorithm}
Algorithm \ref{algo::mppi} shows the proposed framework and is also illustrated in Fig. \ref{fig::main_fig}. 
Let $\s^m_t, w^m_t$ be the sigma points and weights of the $m^{th}$ sampled trajectory at time $t$. Further, let $\delta_{j,i,t}^m$ be the binary switching variable corresponding to mode $j$ of the sigma point $\s_{t, i}^m$ corresponding to $m^{th}$ sample.

In lines 2 and 3, we sample the perturbations and compute the control input for each sample. The timestamped robot trajectory, for 3 samples, is illustrated in Fig. \ref{fig::main_fig} with the green, violet, and orange curves. Line 4 loops over the $M$ sampled control perturbed control input sequences. For each control input sequence, line 7 loops over the time horizon and predicts future states. At each time $t$, Line 8 computes the cost incurred at the current state, lines 9-17 determine the operation mode of the human, and line 18 computes the human sigma points at the next time step. When the robot is outside the attention zone, described by the field-of-view in our example, of the human, the human dynamics is unaffected by the robot, shown by the human states for the first five-time steps in Fig. \ref{fig::main_fig}. Upon perceiving the robot in its field-of-view, which happens for first time at $t=6$ in Fig. \ref{fig::main_fig}, each sigma point of the human determines its operation mode in lines 12 and 14 and continues propagating sigma points to future.

\begin{algorithm}[h!]
\caption{ECUT-MPPI Controller}
\begin{algorithmic}[1]
    \Require $(\mu_0, \Sigma_0)$, $x_{r,0}$, $H$, $F$ \Comment{current human state moments, current robot state, time horizon, Stochastic Dynamics function}
    \State $\s_{0}, w_0\leftarrow $ generate\_UT\_points( $\mu_0, \Sigma_0$ )
    \State Sample $\bm{\epsilon}^m = [\epsilon^m_1, .., \epsilon^m_H]\sim \mathcal{N}(0,\Sigma_\epsilon), \forall m\in \{1,..,M\}$
    \State $\mathbf{u}^m\leftarrow \mathbf{v} + \bm{\epsilon}^m$ 
    \For {$m=1$ to $M$} \Comment{Loop over samples}
        \State $\s_{0}^m, w^m_0 \leftarrow \s_0, w_0$, $x^m_{r,0}\leftarrow x_{r,0}$ 
        \State $S^m = 0$ \Comment{Initialize Cost}
        \For {$t=0$ to $H$} \Comment{Loop over time horizon}
            \State $S^m = S^m + Q(x^m_t) + \gamma (v^m_t)^T \Sigma_\epsilon^{-1} (u^m_t)$
            \For {all $j\in \{1,..,N\}$} \Comment{Operation Modes}
            \For {all $\s_{t,i}^m \in \s_{t}^m$} \Comment{All Sigma points}
            \If{$g(x^m_r, \s_{t,j})\leq 0$} 
            \Statex \quad \quad \Comment{\hfill Determine Switching variable}
                \State $\delta_{j,i,t}^m=1$
            \Else 
            \State $\delta_{j,i,t}^m=0$
            \EndIf
            \EndFor
            \EndFor
            \State $\s_{t+1}^m, w_{t+1}^m\leftarrow$ ECUT$(F(\delta_{i,j,t}^m), \s_t^m, w_t^m )$
            \Statex \Comment{\hfill Get sigma points of next state Algorithm \ref{algo::ec-ut}}
            \State $x^m_{r,t+1}\rightarrow f_r( x_{r,t}^m, u^m_{r,t} )$            
        \EndFor
    \EndFor
    \State $w^m \leftarrow $ \eqref{eq::mppi_weight}
    \State $\mathbf{v}^+ \leftarrow $\eqref{eq::mppi_final_control}
    \Return $\mathbf{v}^+$
\end{algorithmic}
\label{algo::mppi}
\end{algorithm}

\section{Simulation Results}

In this section, we present two case studies to showcase our proposed framework. Code and videos available at \href{https://github.com/hardikparwana/social-navigation}{https://github.com/hardikparwana/social-navigation}.

\subsection{Multi-agent navigation}
\label{section::simulation_case_1}
\begin{figure}
    \centering
    \includegraphics[width=0.45\textwidth]{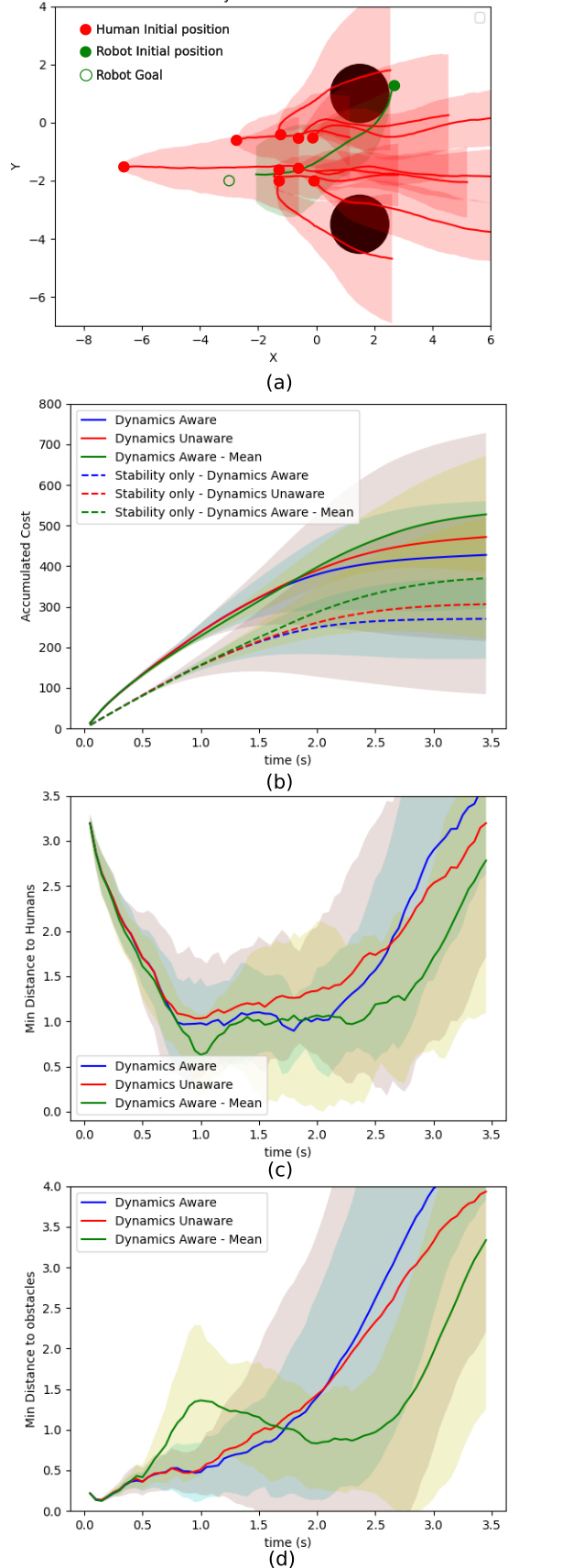}
    \caption{\small{Simulation results for proposed method in Section \ref{section::simulation_case_1}. (a) The robot performs collision avoidance with humans (red) and obstacles (black). Robot trajectory is similarly shown in green. (b) Cost \eqref{eq::mppi_objective} and stability cost ($Q_c$ only) accumulated with time. (c), (d) Robot's minimum distance to any (c) human and (d) obstacle with time. The solid/dotted lines and shaded areas represent the mean and the 95\% confidence interval in variation 
    over 50 runs. }}
    \label{fig::proposed_stats}
    \vspace{-3mm}
\end{figure}

We evaluate the performance of our framework for short-term planning for a robot navigating amongst 10 non-ego agents and 2 circular obstacles. The scenario is shown in Fig \ref{fig::proposed_stats}(a). The robot is modeled as a single integrator with the following dynamics
\eqn{
\dot r_x = u_1, \dot r_y = u_2
}
where $x_r =[r_x ~ r_y]^T$ are the $X,Y$ position of the robot and $u_1,u_2$ are the velocity control inputs. 
 The states of the non-ego agents (other robots or humans for example) are given by $x_p = [p_x ~ p_y]^T$ where $p_x, p_y$ are the position and the control inputs are given by velocities $u_1, u_2$. Every non-ego agent has a sensing radius that defines the switching criteria and thus the two operation modes: cooperative (when the robot is inside the sensing region) and uncooperative (otherwise). The non-ego agents use standard infinity going potential field (PF) for collision avoidance. Since PF is a deterministic model and actual non-ego motions in real life may differ from it, we augment it with the following disturbance model.

\eqn{
   u = u_{nom} + PF(x_r, x_p, x_o) + \mathcal{N}\left(0, \alpha \tanh\left( \frac{\beta}{||[v_x~v_y]||} \right)\right)
   \label{eq::SFM_disturbance}
}
where $u_{nom}=[3,0]^T$ is the nominal velocity of non-ego agents, $PF(x_r, x_p, x_o)$ is the input from PF, $\mathcal{N}(0,\Sigma)$ is a Gaussian distribution with zero mean and $\Sigma$ covariance. The covariance is designed to be inversely proportional to the velocity of the non-ego agent, and $\alpha,\beta>0$ are user-chosen parameters. We designed this model to mimic real-life scenarios where humans and other robots typically change their direction of motion at slower rather than faster speeds.

We compare our algorithm to a baseline based on Risk Aware MPPI \cite{yin2023risk}. We choose the risk metric proposed in \eqref{eq::used_risk} instead of CVaR that \cite{yin2023risk} exclusively uses since centralized confidence interval is easy to compute for both Monte Carlo and UT whereas CVaR estimation using UT is a relatively new formulation\cite{hakobyan2023distributionally} and we refrain from using it in this work for simplicity. The method of risk evaluation is the same as \cite{yin2023risk} - 
for every $m^{th}$ sampled control trajectory of the robot, we simulate $K$ non-ego agent state trajectories where for each $k^{th}_m$ trajectory, the non-ego agent motion is predicted by randomly sampling from its stochastic dynamics function resulting in a total of $MK$ sampled trajectories. The risk for $m^{th}$ robot control sample is thus evaluated using the corresponding $K$ human trajectories. Note that treating each sigma-point as an independent sample, our method equivalently samples a total of $MN$ trajectories where $N=2n+1$ for vanilla UT. In practice, $N<<M$ for similar accuracy \cite{ebeigbe2021generalized, angrisani2005unscented} and therefore our objective here is to compare the achieved risk for similar computational speed.

Our MPPI code is written in JAX and the simulations are performed on a laptop running Ubuntu 22.04 with i9-13905H CPU.
Our code performs in real-time on CPU with up to $M=500$ samples and on GPU with up to $M=20000$ samples. The prediction horizon is $H=40$ with a time step of $\Delta t = 0.05s$. We also choose $\Sigma_\epsilon=4.0$ in \eqref{eq::MPPI-disturbance}, $\alpha=4.0/\Delta t, \beta = 1.0$. We compare three cases: hybrid-dynamics aware, hybrid dynamics unaware, and mean-based prediction for $M=500$ on CPU. 
In the aware case, the robot incorporates a non-ego agent's hybrid motion model in its prediction, whereas in the unaware case, it assumes that non-ego agents are always uncooperative. In the mean-based prediction, the robot performs switching for non-ego agent dynamics based only on their mean positions.
We show that the aware case leads to lower costs while still maintaining safety requirements whereas the mean-based prediction leads to worse results compared to the dynamics aware case. 
We move the humans according to \eqref{eq::SFM_disturbance} and do a Monte Carlo evaluation of our framework by running it 50 times. 
Animations for the simulation can be found on our \href{https://github.com/hardikparwana/social-navigation}{website}.
Figure \ref{fig::proposed_stats}(b) shows the cumulative cost as a function of time and Figs. \ref{fig::proposed_stats}(c) and \ref{fig::proposed_stats}(d) show the nearest distance to any non-ego agent and obstacle at a given time. 
Note that non-ego agents move independently in a decentralized fashion. The dynamics-aware MPPI that predicts non-ego agents' motion based on the knowledge of decentralized dynamics incurs lower costs and observes no constraint violation up to a 95\% confidence interval of MC simulations.
The mean-based prediction (green curves in Fig. \ref{fig::proposed_stats}) and dynamics-unaware case (red curves) initially perform well but the performance degrades with time - the cost increases and constraints are violated (distance to humans is 0 and to obstacles boundary less than 0).

\begin{figure}
    \centering
    \includegraphics[width=0.45\textwidth]{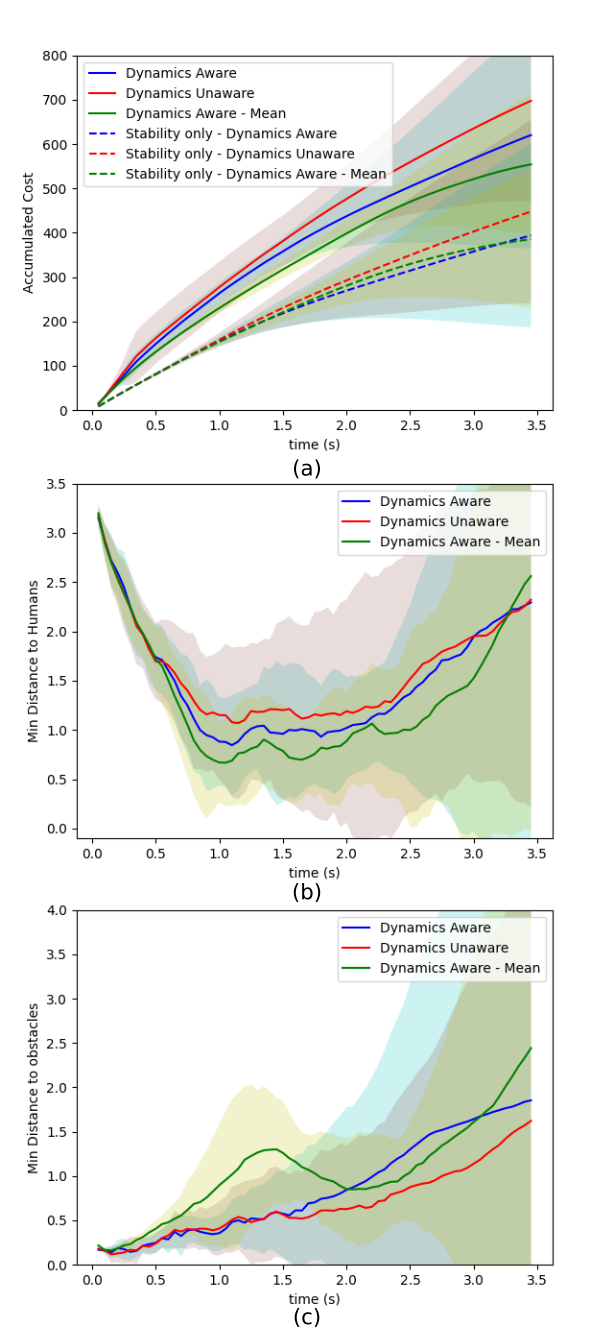}
    \caption{\small{Simulation results for RA-MPPI inspired method in Section \ref{section::simulation_case_1}. Note from the increasing stability cost that the robot didn't converge to the goal location in the given time.}}
    \label{fig::ramppi_stats}
    \vspace{-4mm}
\end{figure}

\begin{figure*}
    \centering
    \vspace{-2.5mm}
    \includegraphics[width=1.0\textwidth]{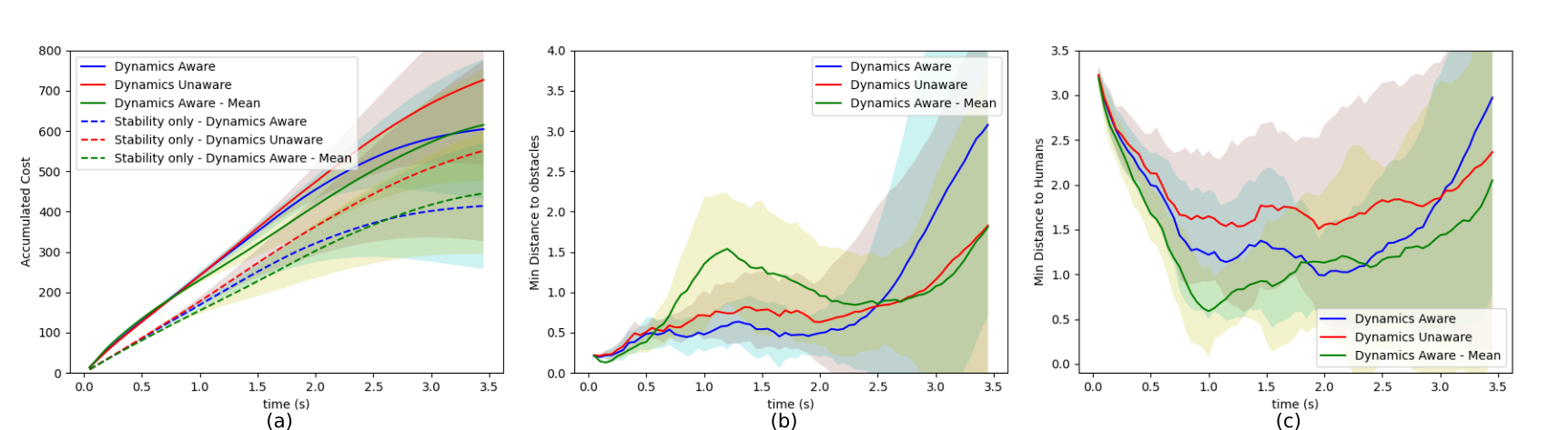}
    \caption{\small{Simulation results for RA-MPPI with 100 samples per human in Section \ref{section::simulation_case_1}. }}
    \label{fig::ramppi_100stats}
    \vspace{-5mm}
\end{figure*}

For RA-MPPI inspired method, we find that choosing $K=20$ leads to a similar computation time per MPPI iteration as our method, and show its results in Fig. \ref{fig::ramppi_stats}. We find that RA-MPPI insp[tired method fails to satisfy safety with the desired probabilistic bounds. For $K>20$, the RA-MPPI inspired method takes longer than our method but to compare purely in terms of performance, we also show results for $K=100$ (which runs 10 times slower than our algorithm) in Fig. \ref{fig::ramppi_100stats}. Note that $K=200$ is much closer to our results (although still has violations) suggesting that the accuracy of UT-based prediction in our method is equivalent to predicting using large MC samples, which is expected.  

We want to clarify that the implementation speed depends significantly on the programming language and hardware acceleration. Our claims on the number of particles required for similar performance between the inspired method and the proposed method are only representative of the behavior we observed in our implementation. Also, note that while the proposed method works better in our case study, Monte Carlo is more accurate than existing UT algorithms when a large number of particles are chosen. Monte Carlo is also better at representing multi-modal distributions than existing UT algorithms. Nevertheless, we reiterate that our method is suitable for real-time implementation where hardware limitations impose constraints on the number of particles. Moreover, our algorithms can always incorporate any improvements in UT algorithms. Finally, note that each non-ego agent is 2-dimensional, and therefore the total size of the state space is 22 and the computation time is expected to increase drastically despite software optimizations. 

\subsection{AWS Hospital Gazebo Experiments}
We evaluate our framework on a gazebo simulator moving the robot in AWS Hospital environment that was developed in our previous work \cite{parwana2023feasible}. We run our JAX code on an NVIDIA RTX 4060 Laptop GPU with 10,000 samples. 

The humans are modeled using a social force model\cite{minami2023development} standard distance-based potential field (PF) to simulate more realistic motions than the standard potential field. Our algorithm though uses the model in \eqref{eq::SFM_disturbance} to predict humans' trajectories as SFM uses a 4-dimensional state per human which severely constrains the number of samples used by MPPI. We also limit trajectory prediction to the nearest 4 humans rather than all the humans. This results in our algorithm being run at 20 Hz.

The robot follows the unicycle model given by following equations $\dot r_x = u_1 \cos r_\theta, \dot r_y = u_1 \sin r_\theta, \dot r_\theta = u_2$
where $r_x,r_y,r_\theta$ are the position and heading angle of the robot and $u_1,u_2$ are the linear and angular velocity control input. We use the ROS2 navigation stack's planner to plan a global path to the goal at the beginning and the proposed MPPI framework as a receding horizon controller to track the global path's waypoints. The video can be found on our \href{https://github.com/hardikparwana/social-navigation}{GitHub} website.

\section{Conclusion}
We proposed a variant of MPPI using Unscented Transform for stochastic hybrid dynamical systems. Our future state prediction scheme can map the uncertainty of states to the uncertainty in switching. We are also able to implement our framework in real-time on both CPU and GPU subject to limits on the number of samples. Future work includes experimentation on real robots and evaluation of different types of hybrid dynamical systems.

\bibliographystyle{IEEEtran}
\bibliography{references.bib}

\end{document}